\documentclass{article}

\usepackage[preprint]{neurips_2025}

\usepackage[utf8]{inputenc} 
\usepackage[T1]{fontenc}    
\usepackage{hyperref}       
\usepackage{url}            
\usepackage{booktabs}       
\usepackage{amsfonts}       
\usepackage{nicefrac}       
\usepackage{microtype}      
\usepackage{xcolor}         
\usepackage{graphicx}
\usepackage{subcaption}
\usepackage{adjustbox}
\usepackage{amsmath}

\title{EVO-LRP: Evolutionary Optimization of LRP for
Interpretable Model Explanations}

\author{%
  Emerald Zhang\footnotemark[1] \\
  University of Texas at Austin\\
  \texttt{emerald.zhang@utexas.edu} \\
  \And
  Julian Weaver\thanks{Equal contribution.} \\
  University of Texas at Austin\\
  \texttt{julian.weaver@utexas.edu} \\
    \And
  Samantha R. Santacruz \\
  University of Texas at Austin \\
  \texttt{srsantacruz@utexas.edu} \\
  \And
  Edward Castillo \\
  University of Texas at Austin \\
  \texttt{edward.castillo@utexas.edu} \\
}

\begin{document}

\maketitle

\begin{abstract}
Explainable AI (XAI) methods help identify which image regions influence a
model’s prediction, but often face a trade-off between detail and interpretability.
Layer-wise Relevance Propagation (LRP) offers a model-aware alternative. However,
LRP implementations commonly rely on heuristic rule sets that aren’t optimized for
clarity or alignment with model behavior. We introduce EVO-LRP, a method that
applies Covariance Matrix Adaptation Evolution Strategy (CMA-ES) to tune LRP
hyperparameters based on quantitative interpretability metrics, like faithfulness or
sparseness. EVO-LRP outperforms traditional XAI approaches in both interpretability metric performance and visual coherence, with strong sensitivity to class-specific
features. These findings demonstrate that attribution quality can be systematically
improved through principled, task-specific optimization.
\end{abstract}

\section{Introduction}

Machine learning models are vital for classification and prediction tasks.
Their increasing complexity and scale yield high predictive accuracy.
However, this sophistication often results in "black box" models.
These models offer little insight into their decision-making.
This opacity makes intuitive and quantitative interpretation difficult.
A growing need exists for systems that explain model outputs robustly and interpretably.
Explainable AI (XAI) aims to clarify how complex models make decisions.
XAI methods typically generate attribution maps (or heat maps).
These maps highlight input portions that most influenced a model's output \cite{devireddy2025comparativestudyexplainableai, xaioverview}.
Many current techniques, like Integrated Gradients (IG) \cite{sundararajan2017axiomaticattributiondeepnetworks}, are model-agnostic.
Model-agnostic methods do not leverage a model's internal architecture.
Consequently, they may fail to produce class-specific explanations.
Sometimes, they return the same visualization regardless of the target class.

Model-dependent approaches atttempt to address these drawbacks.
Layer-wise Relevance Propagation (LRP) is one such method \cite{lrpoverview}.
LRP propagates predictions backward through the model.
It distributes relevance scores across layers using local propagation rules.
However, current LRP implementations often rely on heuristic rule sets and hyperparameters.
These defaults may not be optimal for every model or task.

Generating explanations is only part of the problem.
Evaluating their quality is also essential.
Heatmaps offer visual interpretability- the human eye can assess if highlighted features seem to correspond to the correct class.
However, visual intuition is insufficient for many practical applications.
Without an objective standard, determining if an attribution truly reflects model reasoning is difficult.
It is also hard to know if the explanation is accurate or aligns with the model's decision process.
Subjective assessment alone is inadequate for benchmarking XAI methods.
More rigorous evaluation is required for robust comparisons.
Many evaluation strategies exist but are not standardized across studies.
We rely on well-established benchmarks to avoid bias and complexity.
These benchmarks capture multiple perspectives on explanation quality like faithfulness, sparseness, and robustness \cite{bhatt2020evaluatingaggregatingfeaturebasedmodel, hedstrom2023quantusexplainableaitoolkit}.

We propose EVO-LRP- a method for generating high-fidelity, interpretable attributions.
It applies Covariance Matrix Adaptation Evolution Strategy (CMA-ES) to tune LRP hyperparameters \cite{hansen2023cmaevolutionstrategytutorial}.
This evolutionary strategy tunes relevance propagation rules.
The tuning is based on objective explainability criteria such as faithfulness, sparseness, and sensitivity.
Aligning optimization with these metrics creates explanations that are easier to interpret.
These explanations also seem to align more closely with the decision process of the model.
Our results show that algorithmic parameter tuning improves attribution maps' visual quality.
It also improves their performance on quantitative benchmarks.
Furthermore, EVO-LRP is sensitive to different prediction class.
It produces distinct and focused explanations in multi-class settings.
This demonstrates that attribution quality can be systematically improved through principled, task-specific optimization.

\section{Related Work}
Many methods exist to generate heatmaps.
These heatmaps highlight input regions important for a model’s prediction.
Most methods fall under two broad taxonomies: model-agnostic and model-specific techniques \cite{devireddy2025comparativestudyexplainableai}.

Model-agnostic methods operate independently of the underlying architecture.
This independence makes them flexible and widely applicable across machine learning models.
However, they often require post hoc analysis \cite{Budhkar2025XAI}.
They may also need many forward passes, which increases computational cost.
In contrast, model-specific methods use the internal structure of a network.
For example, they use the weights and activations of a convolutional neural network.
This allows them to provide more targeted explanations.
These architecture-aware methods are often more efficient.
They are also typically more aligned with the model’s behavior.

Within each of these two categories, we can further group XAI methods.
The grouping depends on how they compute relevance: gradient-based or attribution-based.

Gradient-based methods assume access to a model’s parameters \cite{wang2024gradientbasedfeatureattribution}.
They use backpropagation to compute feature relevance.
This involves tracing gradients from the output back to the input.
By doing so, they estimate how changes in each input feature would affect the output.
A key advantage is efficiency.
A single forward and backward pass often yields complete attribution maps.
Integrated Gradients (IG) is a widely used gradient-based method \cite{sundararajan2017axiomaticattributiondeepnetworks}.
IG multiplies averaged gradients by the input features to produce a final attribution score.
Although IG satisfies useful theoretical properties like sensitivity and implementation invariance, its explanations are often class-agnostic in practice \cite{Nielsen_2022}.
They do not always reflect differences between competing class logits.
Class Activation Maps (CAMs) \cite{zhou2015learningdeepfeaturesdiscriminative}, including GradCAM \cite{Selvaraju_2019}, offer a class-specific alternative.
GradCAM uses the gradients of the target class to weight the final convolutional feature maps.
This process produces a heatmap that highlights class-discriminative regions.
GradCAM is frequently used in applications such as weakly-supervised semantic segmentation.
However, the resolution of its output is limited by the spatial size of the final convolutional layer.
Thus, GradCAM attributions are generally upsampled and can be viewed as a mask to the input.

Attribution-based methods offer a different approach \cite{abhishek2022attributionbasedxaimethodscomputer}.
These techniques decompose the model’s output into contributions from each input feature, layer by layer.
Many of these methods build on the Deep Taylor Decomposition (DTD) framework \cite{sixt2022rigorousstudydeeptaylor}.
In this framework, relevance is passed backward through the network.
This uses propagation rules that approximate the local behavior of neurons.
A widely used attribution-based method, and the subject of this paper, is LRP \cite{lrpoverview}.
Other attribution methods in this category include RAP \cite{nam2019relativeattributingpropagationinterpreting}, AGF \cite{gur2020visualizationsupervisedselfsupervisedneural}, DeepLIFT \cite{shrikumar2019learningimportantfeaturespropagating}, and DeepSHAP \cite{lundberg2017unifiedapproachinterpretingmodel}.
Some of these methods, including standard LRP, may exhibit class-agnostic behavior.
This occurs unless they are explicitly designed to be class-aware.
One workaround is to contrast relevance maps for different classes to extract discriminative regions.

Outside of gradient- and attribution-based methods are perturbation-based methods.
LIME (Local Interpretable Model-Agnostic Explanations) is an example \cite{ribeiro2016whyitrustyou}.
LIME perturbs input features and observes the resulting change in model output.
It then fits a simple surrogate model- typically linear- to estimate feature importance.
This approach does not require gradients or access to model internals.
This makes it compatible with any classifier.
However, it often produces coarse and unstable explanations.
The resulting surrogate model may not generalize well.
Because LIME perturbs only a subset of input features, it must run many forward passes.
This increases computational cost.
Other perturbation-based methods, while interpretable, face similar scalability issues.
They also lack fine-grained spatial precision \cite{ribeiro2024reliablestableexplanationsxai, Salih_2024}.
We include LIME as a benchmark in our evaluation.
It represents a widely used, architecture-agnostic baseline.

\subsection{Layer-wise Relevance Propagation (LRP)}

Although LRP is often categorized as model-agnostic, it uses the network’s computational graph.
This graph is used to back-propagate relevance.
This allows it to generate detailed explanations quickly.
The LRP framework preserves a conservation property.
This property means the total relevance at each layer matches the output prediction.
It ensures consistency across layers.
It also avoids the numerical instability seen in some gradient-based techniques.

Researchers have proposed various heuristics for assigning LRP rules across different network layers.
For example, in VGG16, the upper layers tend to represent overlapping class concepts \cite{lrpoverview}.
In these layers, LRP-0 is thought to perform well.
It tracks the raw activation pattern and its gradient.
In the middle layers, where multiple filters are stacked, LRP-$\epsilon$ can be used \cite{lrpoverview}.
LRP-$\epsilon$ can suppress irrelevant variation and focus relevance on the most salient features.
These intuitions are based on theoretical reasoning.
However, they have not been rigorously validated in practice.

Furthermore, there have been prior efforts at hyperparameter selection in LRP \cite{pahde2023optimizingexplanationsnetworkcanonization, achtibat2024attnlrpattentionawarelayerwiserelevance} that utilize grid-search as the fundamental algorithm for tuning and similar XAI metrics as quantitative evaluation. These endeavors have resulted in significant improvement in heatmap quality, but experiments are often limited to a single type of LRP rule and a discrete set of possible hyperparameter choices. 

Despite LRP’s flexibility, current implementations rarely use composite rule sets.
Rules like LRP-$\alpha\beta$, which allow more expressive tuning, are underused.
Their underutilization is due to the complexity of selecting their parameters.
This significantly limits the method’s potential.
It may lead to noisy or diffuse heatmaps when LRP is applied out-of-the-box.

\subsection{Optimization for Explainability}

Evolutionary Algorithms (EAs) are a class of optimization methods.
They are inspired by natural selection.
These algorithms solve complex problems by maintaining a population of candidate solutions.
Over time, they apply mutation, recombination, and selection to evolve better-performing solutions.
EAs are widely used in tasks where exact solutions are intractable.
They are also used when the objective is noisy and non-differentiable.

A prominent EA for continuous optimization is Covariance Matrix Adaptation Evolution Strategy (CMA-ES) \cite{hansen2023cmaevolutionstrategytutorial}.
CMA-ES operates by sampling candidates from a multivariate Gaussian distribution.
It updates this distribution over generations by adjusting the covariance matrix.
This enables CMA-ES to learn correlations between parameters.
It can also efficiently search ill-conditioned landscapes.

Researchers have recently applied evolutionary computation to interpretability \cite{zhou2024evolutionarycomputationexplainableai}.
For example, Genetic Programming Explainer (GPX) evolves symbolic expression trees \cite{ferreira2020applyinggeneticprogrammingimprove}.
These trees locally approximate a model’s decision around a given input.
Like LIME, GPX samples neighboring points and fits a local surrogate.
But unlike LIME’s linear model, GPX uses genetic programming to evolve more expressive forms.
Other work has used genetic algorithms to optimize robustness metrics \cite{huang2023safariversatileefficientevaluations}.
Some have used them to generate adversarial examples for stress-testing XAI methods \cite{Sharma_2020, Qiu_2021}.

Despite these advances, to our knowledge, no prior work has applied evolutionary optimization directly to the rule selection and parameter tuning of LRP.
This represents a missed opportunity.
LRP includes multiple rules and tunable parameters.
These strongly influence attribution quality.
A principled, data-driven approach to rule selection, like the one we introduce with EVO-LRP, can unlock LRP’s full flexibility.
It can also improve alignment with interpretability metrics.

\subsection{Quantitative Evaluation Metrics}

Evaluating explanation quality remains a central challenge in XAI \cite{Nauta_2023}.
While many studies rely on visual inspection, quantitative metrics provide a more objective basis.
They allow for comparing attribution methods.
Tools such as Captum \cite{kokhlikyan2020captumunifiedgenericmodel}, iNNvestigate \cite{alber2018innvestigateneuralnetworks}, and Quantus \cite{hedstrom2023quantusexplainableaitoolkit} implement widely used evaluation metrics.
These include faithfulness, sparseness, and sensitivity.
These metrics assess how accurately an explanation reflects the model’s reasoning.
They also assess how concentrated the attribution is on relevant features.
Additionally, they measure how stable the explanation remains under small perturbations.
However, these metrics are often used inconsistently.
Their definitions can vary across studies.
The metrics we use- faithfulness, sparseness, and sensitivity- are straightforward to compute for a given explanation.
Crucially, they are non-differentiable with respect to the LRP parameters.
This makes them a natural fit for optimization with CMA-ES.
Implementation details are described in the Methodology section.

\section{Methodology}

Our method uses Layer-wise Relevance Propagation (LRP) as its base explanation framework.
Given an input and a class label, LRP computes attributions.
It does this by propagating relevance scores backward through the model.
We extend this LRP process.
We optimize the hyperparameters associated with each LRP rule.
This optimization uses the Covariance Matrix Adaptation Evolution Strategy (CMA-ES).
The objective of this optimization is to maximize selected quantitative evaluation metrics.

\subsection{Terminology}

We define four key terms used throughout this paper.
\begin{itemize}
    \item[] \textbf{Rule:} An LRP rule, such as LRP-$0$, LRP-$\varepsilon$, or LRP-$\alpha\beta$.
    \item[] \textbf{Value:} The hyperparameter associated with a given rule (e.g., $\alpha$, $\beta$, or $\varepsilon$).
    \item[] \textbf{Layer:} A layer in the model (e.g., convolutional or linear) to which a rule and value is applied.
    \item[] \textbf{Uniform Rule Optimization (URO):} Assigns the same rule to all layers.
    URO allows the hyperparameter value to vary by layer.
    In URO, the parameter space is the number of trainable layers.
\end{itemize}
Implementations and formulas for the LRP rules can be found in the technical appendix \ref{lrp_rule_impl}.



    
    

\subsection{Optimization Procedure}

We use CMA-ES for optimization.
For single-objective optimization, we use the \texttt{pycma}\cite{hansen2023cmaevolutionstrategytutorial} package.
For bi-objective cases, we use the \texttt{pycomocma}\cite{Tour__2019} library.
These optimizers do not require gradients of the objective function.
Our objective functions are defined by the evaluation metrics described below.

\subsection{Evaluation Metrics}

We select three quantitative metrics for optimization.

\begin{itemize}
    \item \textbf{Faithfulness Correlation (FC)\cite{bhatt2020evaluatingaggregatingfeaturebasedmodel}:}
    FC assesses how well an explanation reflects the model’s actual behavior.
    It measures the correlation between the model’s output changes when certain features are masked.
    This correlation is with the relevance scores assigned to those masked features.
    A higher correlation indicates the explanation accurately captures feature influence on predictions. Faithfulness scores are always positive.
  \[
  \text{FC} = \left| \text{corr}(\text{logits}(x_{\setminus S}), R_S) \right|
  \]  Where $x_{\setminus S}$ denotes input with masked features, $R_S$ is the relevance assigned to $S$. 
    \item \textbf{Average Sensitivity (AS)\cite{yeh2019infidelitysensitivityexplanations, bhatt2020evaluatingaggregatingfeaturebasedmodel}:}
    AS evaluates the robustness of an explanation.
    It examines how much the attribution scores change in response to small, random perturbations in the input.
    If minor input changes lead to significant variations in the explanation, it suggests sensitivity.
    Such sensitivity can indicate potential unreliability.
    Lower average sensitivity values indicate more stable and trustworthy explanations. Average Sensitivity values fall between 0 and 1.
  \[
  \text{AS} = \mathbb{E}_{\delta \sim \mathcal{N}(0, \sigma^2)} \left[ \| R(x + \delta) - R(x) \|_2 \right]
  \]
    \item \textbf{Sparseness (SP) \cite{chalasani2020conciseexplanationsneuralnetworks}:}
    Sparseness measures the conciseness of an explanation.
    It determines how focused the attribution is across the input features.
    A sparse explanation concentrates relevance on a few key features.
    This concentration makes the explanation easier to interpret.
    Higher sparseness values imply a more concise explanation.
    Such an explanation highlights the most influential features without unnecessary complexity. Sparseness takes on a value between 0 and 1. 
  \[
  \text{SP} = \frac{\sum_i |R_i|}{\| R \|_2}
  \]
\end{itemize}

\subsection{Experimental Setup}
We use the pre-trained VGG16\cite{simonyan2015deepconvolutionalnetworkslargescale} network from PyTorch.
We evaluate our method on the ILSVRC 2012 ImageNet \cite{russakovsky2015imagenetlargescalevisual} validation set.
This dataset contains 50,000 images from 1,000 classes.
Many of these images contain objects from multiple classes. The hyperparameter set produced by EVO-LRP is tuned on a random batch of size 64 from ILSVRC 2012. 
We compare EVO-LRP to several commonly used explainability baselines.
These baselines are GradCAM, Integrated Gradients (IG), LIME, and LRP-$0$.
GradCAM is gradient-based and class-specific.
Integrated Gradients is gradient-based and often class-agnostic.
LIME is perturbation-based and model-agnostic.
LRP-$0$ is attribution-based and model-specific.
All baseline methods are implemented using the \texttt{Captum} \cite{kokhlikyan2020captumunifiedgenericmodel} library for consistency.
Evaluation results are reported across all three selected metrics.

\section{Results}

\subsection{Quantitative Evaluation}

We benchmarked EVO-LRP against common attribution techniques: LIME, Integrated Gradients (IG), GradCAM, and standard LRP-$0$.
Performance was evaluated using faithfulness, sparseness (higher is better), and average sensitivity (lower is better).
Experiments used the ImageNet 2012 validation set, a rigorous testbed due to its class diversity and visual complexity.

\begin{table}[ht]
\centering
\caption{Attribution evaluation scores for baseline methods. Arrows indicate preferred direction ($\uparrow$ for higher, $\downarrow$ for lower). Bold values denote the best performance per metric. LIME*'s high faithfulness has high variance. Standard deviations are calculated over the individual samples in the validation split.}
\vspace{\baselineskip}
\label{baselines}
  \begin{adjustbox}{width=\textwidth}
\begin{tabular}{lcccccccc}
\toprule
 & \multicolumn{2}{c}{\textbf{LIME}} & \multicolumn{2}{c}{\textbf{Integrated Gradients}} & \multicolumn{2}{c}{\textbf{GradCAM}} & \multicolumn{2}{c}{\textbf{LRP-0}} \\
 & Mean & Std & Mean & Std & Mean & Std & Mean & Std \\
\midrule
\textbf{Faithfulness$\uparrow$}    & \textbf{2.24E+00*} & 2.07E+00 & 7.03E-02 & 1.76E-02 & 3.55E-01 & 7.35E-02 & 8.57E-02 & 1.18E-02 \\
\textbf{Sparseness$\uparrow$}        & 4.81E-01 & 6.97E-02 & 4.81E-01 & 3.08E-02 & 2.45E-01 & 2.03E-02 & \textbf{4.92E-01} & 1.47E-02 \\
\textbf{Avg Sensitivity$\downarrow$}  & 6.16E-01 & 1.70E-01 & 6.62E-01 & 7.42E-02 & \textbf{4.15E-01} & 5.10E-02 & 1.16E+00 & 1.78E-02 \\
\bottomrule
\end{tabular}
\end{adjustbox}
\end{table}

Table~\ref{baselines} shows baseline method limitations.
LIME achieved the highest mean faithfulness but showed inconsistent behavior (high standard deviation).
IG and GradCAM were more stable but had poor faithfulness scores.
GradCAM performed relatively well on sensitivity but struggled with sparseness and semantic alignment.
LRP-$0$ had reasonable sparseness but high average sensitivity, indicating low robustness.
No baseline approach successfully balanced all three objectives, showing the need for more consistently high-quality explanation methods.

\subsection{Uniform Rule Optimization Improves Relevance Attribution}

We applied single-objective CMA-ES for Uniform Rule Optimization (URO).
This optimized individual LRP value parameters for each trainable layer, adapting relevance propagation layer-by-layer.
We confined the search for each hyperparameter to a valid range.

\begin{table}[ht]
\centering
\caption{EVO-LRP URO results using CMA-ES for LRP-$\alpha\beta$ and LRP-$\epsilon$ rules. Arrows indicate preferred direction. Bold values denote best performance per metric/rule. Standard deviations are calculated over the individual samples in the validation split.}
\vspace{\baselineskip}
\label{uro_results}
\begin{tabular}{lcccc}
\toprule
 & \multicolumn{2}{c}{\textbf{LRP-$\alpha\beta$}} & \multicolumn{2}{c}{\textbf{LRP-$\epsilon$}} \\
 \cmidrule(r){2-5}
 & Mean & Std & Mean & Std \\
\midrule
\textbf{Faithfulness$\uparrow$}    & \textbf{1.46E+00} & 1.10E-01 & 5.78E-02 & 7.92E-03 \\
\textbf{Sparseness$\uparrow$}      & 6.70E-01 & 1.63E-02 & \textbf{7.23E-01} & 1.64E-02 \\
\textbf{Avg Sensitivity$\downarrow$} & \textbf{2.73E-01} & 2.07E-02 & 8.52E-01 & 4.99E-02 \\
\bottomrule
\end{tabular}
\end{table}

\begin{figure}[htbp]
    \centering
    \begin{subfigure}{0.48\textwidth}
        \centering
        \includegraphics[width=\linewidth]{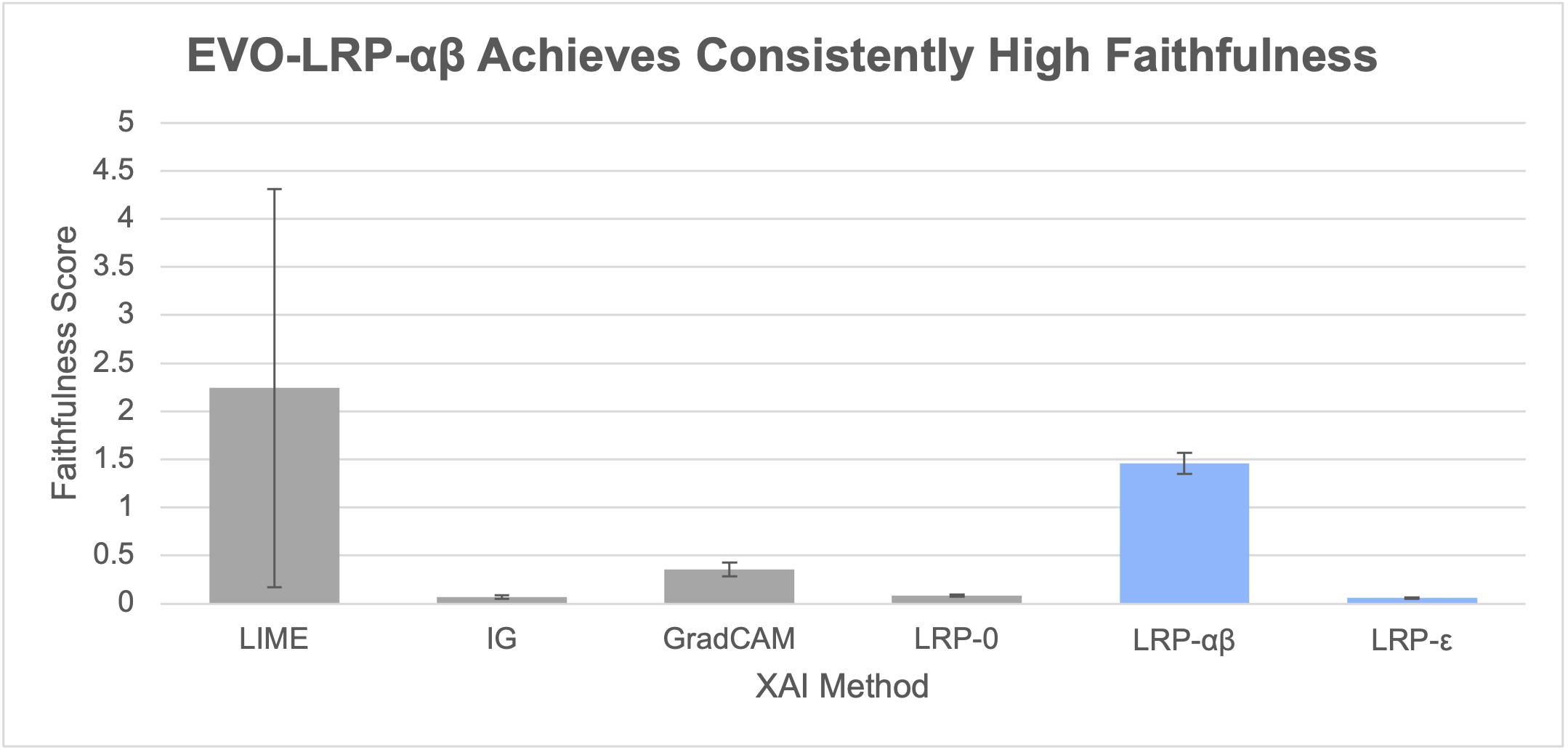}
        \caption{Faithfulness Comparison. EVO-LRP (LRP-$\alpha\beta$) achieves high faithfulness with low variance, significantly outperforming most baselines. This indicates its explanations more reliably reflect the model's true decision process compared to less consistent or less faithful methods.}
        \label{fig:faith_compare}
    \end{subfigure}
    \hfill
    \begin{subfigure}{0.48\textwidth}
    \centering
    \includegraphics[width=\linewidth]{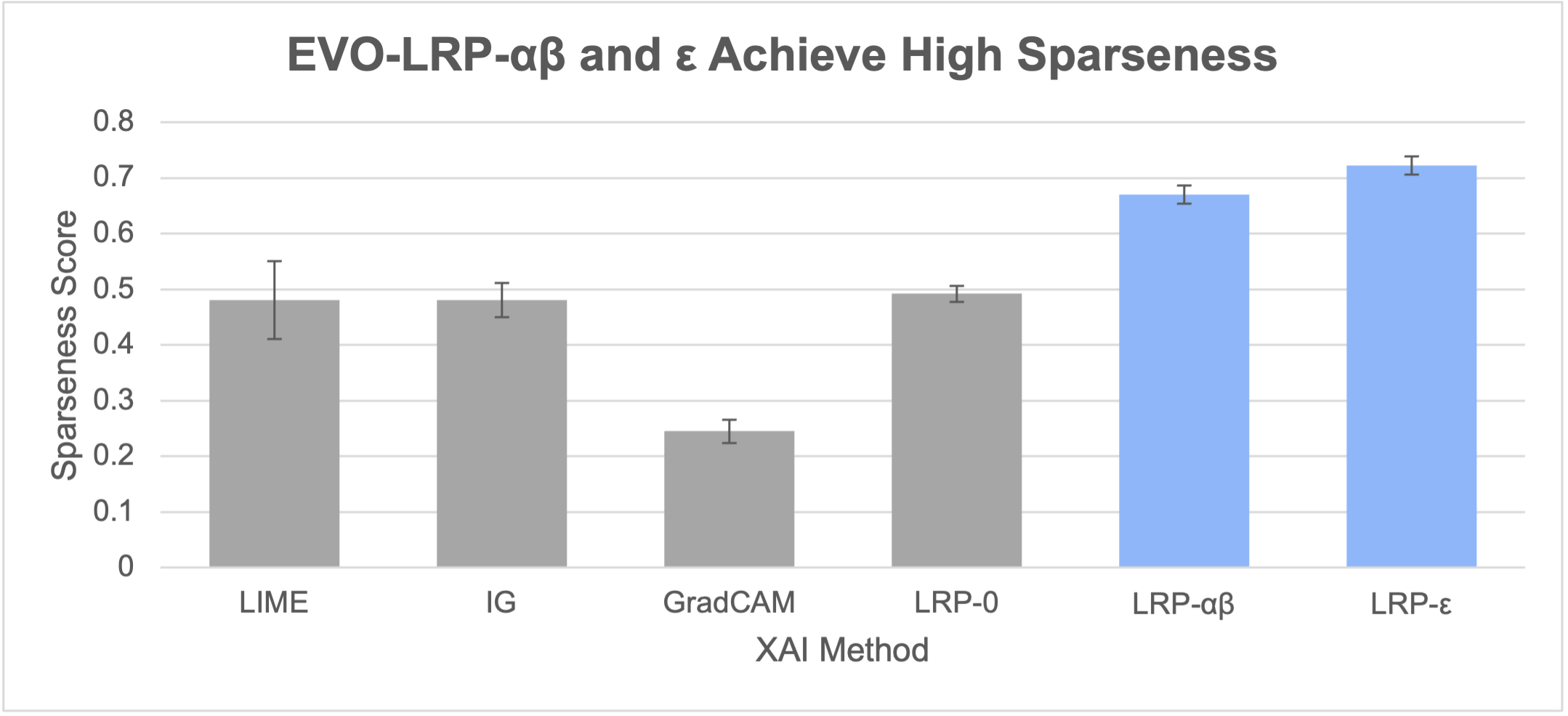}
    \caption{Sparseness Comparison. EVO-LRP (LRP-$\alpha\beta$ and LRP-$\epsilon$) yields significantly sparser (more focused) explanations than all baselines. This enhances interpretability by concentrating relevance on the most critical input features, reducing cognitive load for the user.}
    \label{fig:sparse_compare}
    \end{subfigure}
    \vspace{0.5cm} 
    \begin{subfigure}{0.48\textwidth}
    \centering
    \includegraphics[width=\linewidth]{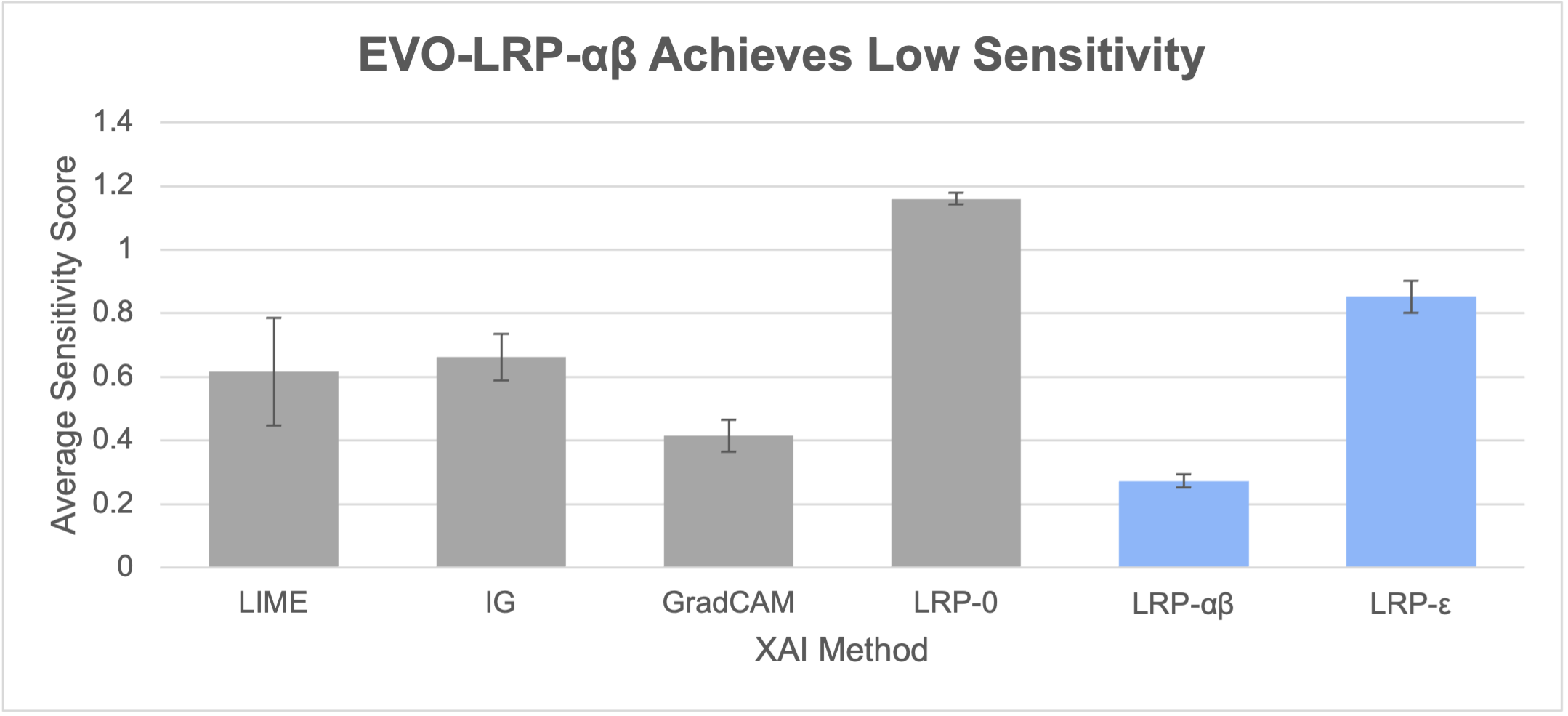}
    \caption{Average Sensitivity Comparison. EVO-LRP (LRP-$\alpha\beta$) shows the lowest average sensitivity, signifying superior explanation robustness. Its stability against minor input perturbations builds trust in the explanation's reliability.}
    \label{fig:sens_compare}
    \end{subfigure}    
    \caption{Quantitative benchmark comparison of EVO-LRP (URO LRP-$\alpha\beta$ and LRP-$\epsilon$) against baselines across key XAI metrics. Error bars: standard deviation. While EVO-LRP systematically optimizes LRP-$\epsilon$ to high sparsity, LRP-$\alpha\beta$ achieves a superior balance of high faithfulness, high sparseness, and low sensitivity. This comprehensive improvement is crucial for generating trustworthy, understandable, and reliable explanations, addressing shortcomings of existing methods.}
    \label{fig:all_method_barchart}
\end{figure}

Table~\ref{uro_results} summarizes results for EVO-LRP with LRP-$\alpha\beta$ and LRP-$\epsilon$ rules.
Figure~\ref{fig:all_method_barchart} quantitatively compares EVO-LRP (URO LRP-$\alpha\beta$) against baselines.

EVO-LRP with the optimized LRP-$\alpha\beta$ rule consistently surpassed baselines across all metrics (Figure~\ref{fig:all_method_barchart}).
Its high faithfulness indicates attributions align with the model’s decision process.
Its low average sensitivity suggests robustness to input perturbations.
This impactfully shows EVO-LRP makes the expressive LRP-$\alpha\beta$ rule practically effective.
Appendix \ref{multiopt} shows similar results from Pareto-based multi-objective optimization.

Optimizing the LRP-$\epsilon$ rule primarily yielded extremely sparse maps (Table~\ref{uro_results}).
It offered limited gains in faithfulness or sensitivity, aligning with LRP-$\epsilon$'s known conservatism.
This can compromise precision and responsiveness.

These findings demonstrate a key contribution: LRP rule choice and systematic parameter tuning are crucial for explanation quality.
EVO-LRP unlocks the LRP-$\alpha\beta$ rule’s potential for more expressive and adaptable attribution.

\subsection{Qualitative Assessment}
\label{singl_opt}
\subsubsection*{Qualitative Assessment of EVO-LRP-\texorpdfstring{$\alpha\beta$}{alphabeta} Reveals Complementary Attribution Patterns}

\begin{figure}[htbp]
    \centering
    \includegraphics[width=0.75\textwidth]{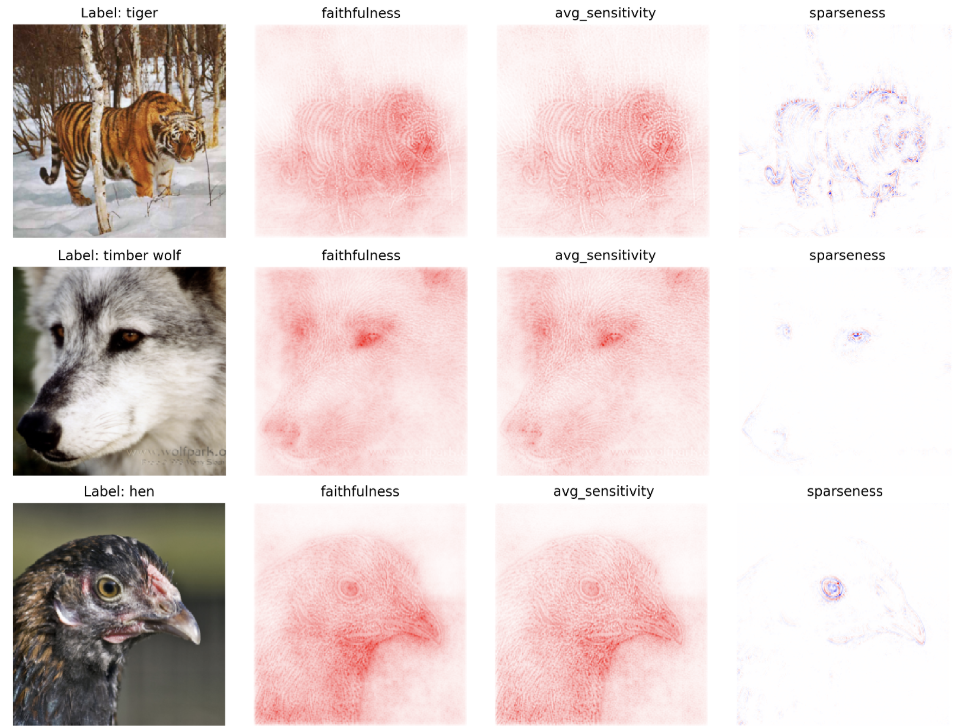}
    \caption{LRP-$\alpha\beta$ all-class relevance maps from EVO-LRP, optimized for individual metrics (faithfulness, avg. sensitivity, sparseness). Faithfulness/sensitivity maps capture broad semantic regions critical to model decisions. The sparseness-optimized map (right) distinctively highlights object boundaries (positive: red, negative: blue). This emergent edge-detection is interesting as it mirrors a fundamental aspect of human visual perception. Edge-detection suggests the model may recognize objects via salient contours, offering insight into its learned features beyond just diffuse areas.}
    \label{fig:UROmetrics}
\end{figure}

We visualized LRP-$\alpha\beta$ relevance maps from models optimized individually for faithfulness, sensitivity, or sparseness.
Figure~\ref{fig:UROmetrics} shows these maps (aggregated across classes).
Faithfulness and sensitivity-optimized models yielded similar maps, emphasizing broad relevant regions.
The sparseness-optimized model consistently highlighted object boundaries with sharp positive (red) and negative (blue) relevance.
This suggests an emergent, valuable property from EVO-LRP's optimization: relevant-edge-detection.

\subsubsection*{Composite Relevance Combines Strengths of Individual Metrics}

\begin{figure}[htbp]
    \centering
    \includegraphics[width=1\textwidth]{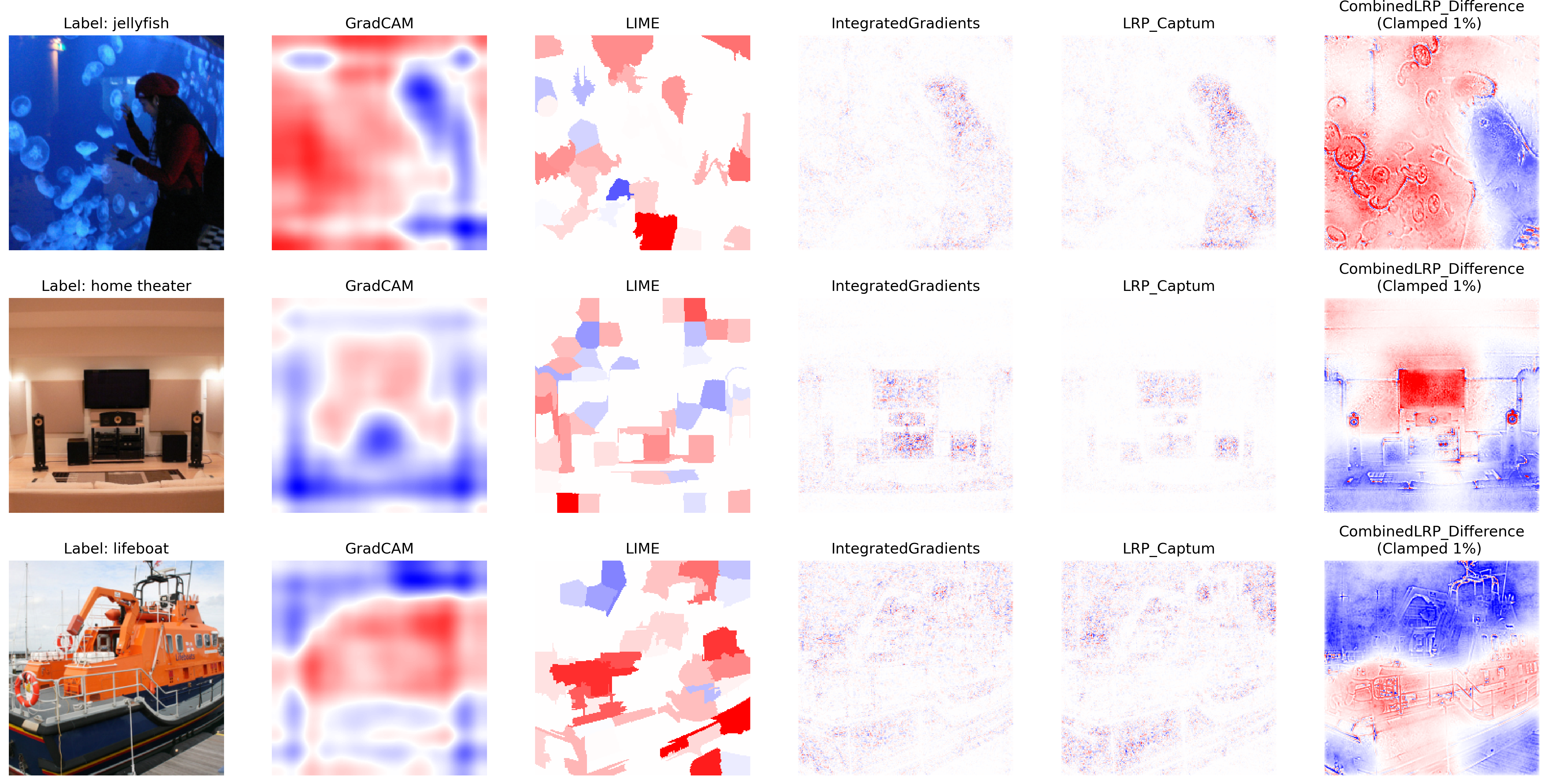}
    \caption{Comparison of relevance maps: Our composite EVO-LRP map (far right) versus standard baselines (GradCAM, LIME, IG, LRP-0). Positive relevance is red, negative blue. We clamped the top and bottom 1\% of relevance scores to minimize the effect of outliers. The composite map, synthesizing strengths from different EVO-LRP optimizations, is visibly sharper, less noisy, and more semantically aligned.}
    \label{fig:compositerelmap}
\end{figure}

We developed a novel method for a composite attribution map, synthesizing strengths from each optimization.
We combined class-specific relevance maps (optimized for faithfulness, sensitivity, sparseness).
To isolate class-specific contributions, we subtracted the all-class relevance map from single-class attributions.
This removes shared background, emphasizing class-unique features.
Summed maps were then clamped (top/bottom 1\% percentile) to suppress sparse map outliers.
The resulting composite map (Figure~\ref{fig:compositerelmap}) retains boundary clarity (from sparseness) and broader semantic structure (from others).
This demonstrates EVO-LRP's practical impact in enabling richer, multi-faceted explanations.

\subsubsection*{Baseline Comparisons Demonstrate Noise and Incoherence}

Compared to our EVO-LRP composite (Figure~\ref{fig:compositerelmap}), baselines were inconsistent in feature selection and spatial precision.
GradCAM often showed diffuse or irrelevant background attributions.
IG maps were noisy and fragmented.
LIME maps were coarse, lacking semantic alignment.
Our composite LRP maps, however, maintained localized, structured, class-specific relevance with minimal noise.
This underscores EVO-LRP's value in optimizing LRP for specific, high-quality interpretability goals.

\subsection{Class-specific Relevance Maps Clearly Distinguish Competing Classes}

\begin{figure}[htbp]
\centering
\begin{subfigure}{0.45\textwidth}
\includegraphics[width=\linewidth]{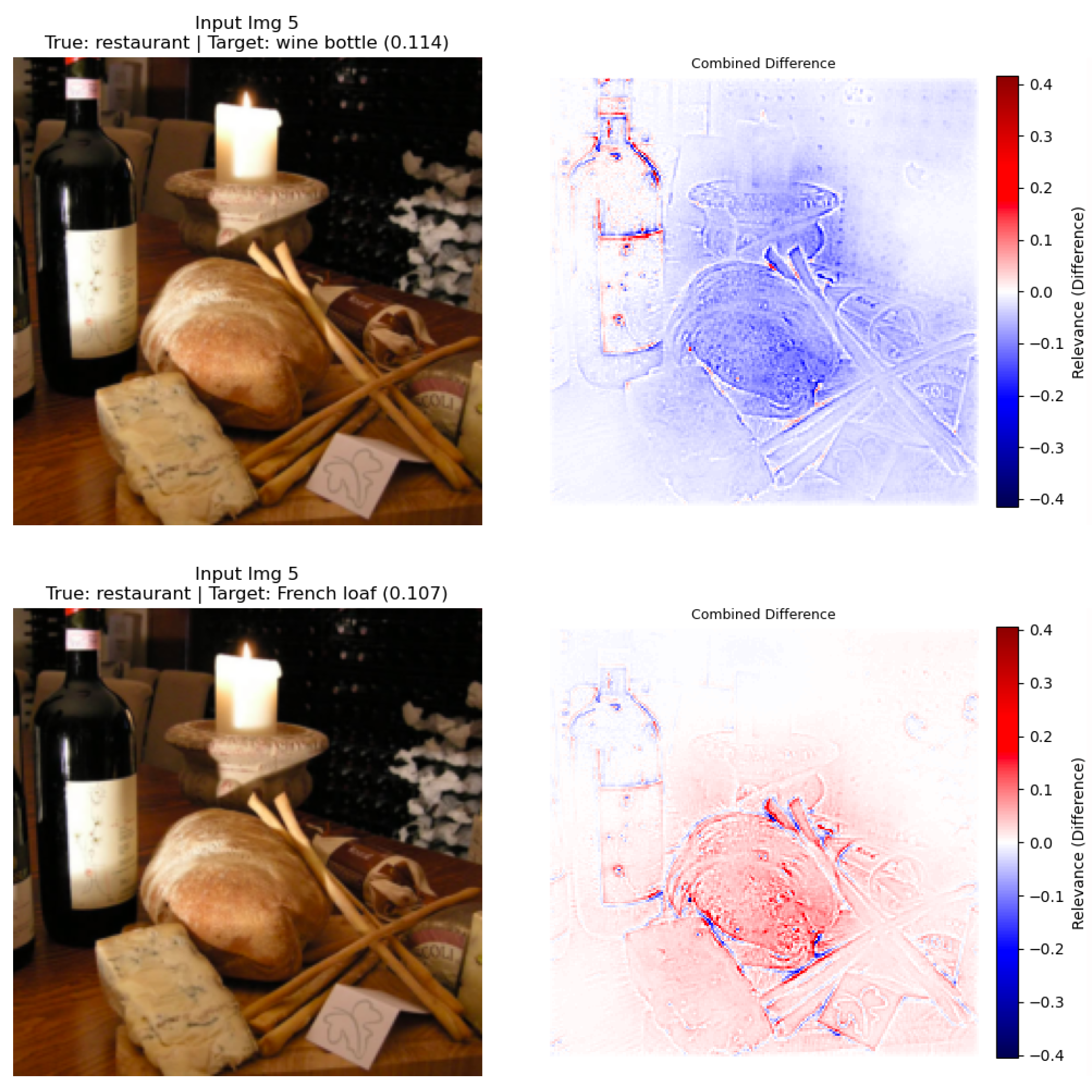}
\caption{EVO-LRP distinguishes \textit{wine bottle} (top) from \textit{french loaf} (bottom). Positive relevance (red) accurately localizes to the target object, while other prominent objects receive negative (blue) relevance or be ignored.}
\label{fig:class_specific_french_wine}
\end{subfigure}
\hfill
\begin{subfigure}{0.45\textwidth}
\includegraphics[width=\linewidth]{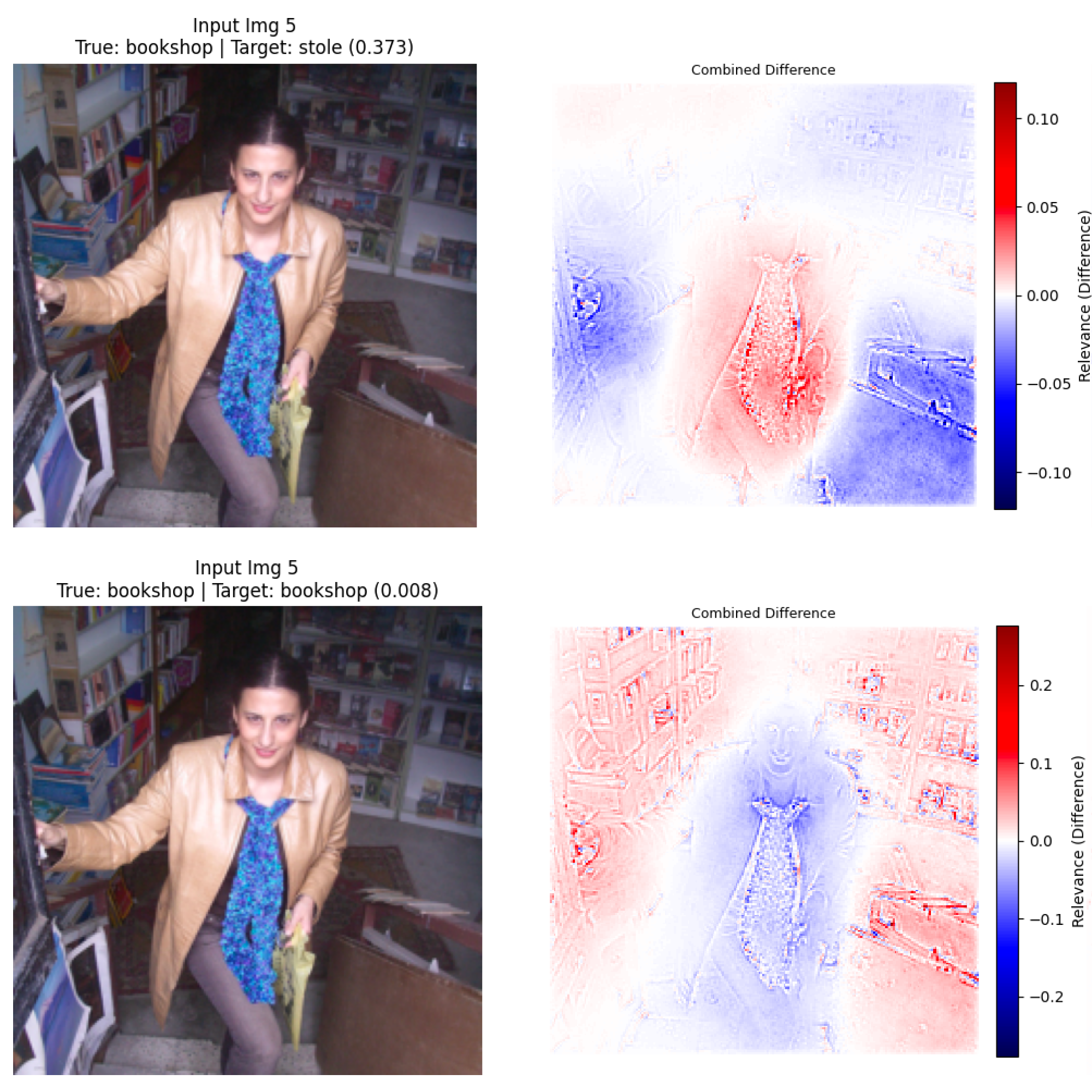}
\caption{EVO-LRP distinguishes \textit{stole} (top) from \textit{bookshop} (bottom). Although the prediction score for the correct class \textit{bookshop} was very low (0.008), EVO-LRP still identifies relevant contextual features, like books in the background.}
\label{fig:class_specific_stole_bookshop}
\end{subfigure}
\caption{EVO-LRP class-specific maps demonstrate clear distinction between competing classes and identification of relevant features. Heatmaps (positive: red, negative: blue) highlight unique semantic features for each target class. This ability to provide targeted, discriminative explanations is vital for understanding why a model chose a specific class over alternatives or what features it associates with a class, crucial for debugging and trust.}
\label{fig:class_specific}
\end{figure}

Crucially, EVO-LRP can generate relevance maps targeted to any class within an image. This provides insight into model decision-making, especially for competing classes.
Figure~\ref{fig:class_specific} shows such class-specific maps, isolating unique class relevance by subtracting average all-class attribution.
Positive relevance is red, negative is blue.

Comparing \textit{french loaf} and \textit{wine bottle} (Figure~\ref{fig:class_specific_french_wine}): the \textit{wine bottle} map emphasized the bottle (bread negative); the \textit{french loaf} map positively highlighted the loaf.
Comparing \textit{stole} and \textit{bookshop} (Figure~\ref{fig:class_specific_stole_bookshop}): the \textit{stole} map highlighted the accessory; the \textit{bookshop} map highlighted shelves and individual books (stole negative).

This nuanced positive/negative feature detection is a novel outcome of EVO-LRP's optimization.
It shows promise for domains like medical imaging needing transparent decision-making.
LIME/IG struggle to distinguish classes this way; GradCAM can, but EVO-LRP offers superior sharpness and less noise.
EVO-LRP maps are clear, high-contrast, with well-defined boundaries, confirming precise, interpretable class-specific attributions.

\subsection{Attribution Structure and Trade-offs}

We observed a trade-off between map sparsity and spatial coverage.
Sparseness optimization yielded edge-detector like maps, highlighting outlines, sometimes ignoring internal regions.
This intriguingly resembles early visual brain processing \cite{huff2025neuroanatomy}.
This is interesting because it suggests the model might be learning perceptually fundamental features (like contours) for object recognition, which can make explanations more intuitive.
Faithfulness/sensitivity optimization highlighted broader, diffuse features capturing overall object shape but with less boundary sharpness.

Baselines showed similar trade-offs.
GradCAM/LIME highlighted large semantic areas but lacked precision.
IG was sparse but often scattered and hard to interpret.
This suggests no single default XAI method captures all desirable explanation qualities.
EVO-LRP allows deliberate exploration of these facets.
Combining perspectives, as with our composite maps, offers a more complete picture of model decisions- a key impact of our flexible framework, as understanding these trade-offs is essential for selecting the most appropriate type of explanation for a given task.

\section{Conclusion}
We have presented EVO-LRP, an optimization framework.
EVO-LRP extends the capabilities of Layer-wise Relevance Propagation (LRP).
It applies Covariance Matrix Adaptation Evolution Strategy (CMA-ES) to tune LRP's rule-specific hyperparameters.
This process improves the alignment between model predictions and their corresponding attributions.
Our experiments demonstrate that automatically optimized LRP parameters yield explanations that are sparse, faithful to the model, and insensitive to minor input perturbations.
EVO-LRP has been formulated as a flexible and generalizable optimization method.
It is not confined to a specific model or dataset.
Furthermore, we designed it to be compatible with other evaluation metrics.
This enables users to adjust the framework to the interpretability goals of a specific use-case.

We believe this work contributes to improving transparency in machine learning systems.
This is particularly important in applications where interpretability is necessary.
For example, in medical domains such as clinical diagnosis or tumor segmentation, models providing clearer explanations may be more readily adopted and trusted.
Future work should focus on applying EVO-LRP to larger-scale architectures, such as vision transformers.
Evaluating it on domain-specific datasets beyond standard image classification tasks is also a priority.
Other avenues of expansion include testing a broader range of evaluation metrics.
Exploring advanced multiobjective optimization strategies will also be pursued.

\label{limitations}
\paragraph{Limitations}
Beyond Uniform Rule Optimization (URO), we explored alternative approaches: Uniform Value Optimization (UVO) and Composite Optimization (CO).
Neither approach consistently outperformed URO.
Composite LRP proved difficult to tune due to its high dimensionality and computational cost.
Assigning rule types across layers requires a complex search over both rule space and network structure.
Uniform Value LRP underperformed unless its chosen fixed rule type happened to align well with the functional role of each layer.
This condition is not reliably met across diverse architectures.
These exploratory variants suggest there is room for innovation, especially in hybrid or adaptive rule selection strategies.
However, they also affirm the strength and practicality of URO as a robust default.
Further constraints or methodological refinements may be needed before UVO or CO can be reliably applied in broader settings.

The impact of EVO-LRP's optimization can also vary depending on the LRP rule type.
Inherently stable rules, such as LRP-$\epsilon$, naturally offer less scope for drastic changes via hyperparameter tuning.
Finally, while EVO-LRP is designed for flexibility, its application to extremely large models or very high-dimensional data (e.g., 4D medical volumes) warrants further investigation.
Specifically, the scalability of compute time in such scenarios must be thoroughly analyzed.

\bibliographystyle{plain}
\bibliography{biblio}

\appendix

\section{Technical Appendices and Supplementary Material}
\label{lrp_rule_impl} 
\subsection{LRP Rule Implementations}

We implement three primary types of LRP rules.
In these rules, $R_k$ represents the relevance at neuron $k$.
$R_j$ is the relevance at a neuron $j$ in a lower layer.
$a_i$ denotes the activation of neuron $i$.
$w_{ij}$ represents the connection weights.

\begin{itemize}
    \item[] \textbf{LRP-$0$:}
    \[
    R_j = \sum_k \frac{a_j w_{jk}}{\sum_{j'} a_{j'} w_{j'k}} R_k
    \]
    This rule redistributes relevance.
    The redistribution is proportional to each input's contribution to neuron $k$'s activation.
    
    \item[] \textbf{LRP-$\varepsilon$:}
    \[
    R_j = \sum_k \frac{a_j w_{jk}}{\sum_{j'} a_{j'} w_{j'k} + \varepsilon \cdot \text{sign}(\sum_{j'} a_{j'} w_{j'k})} R_k
    \]
    This rule adds a stabilizing term, $\varepsilon$, in the denominator.
    The term helps handle small or conflicting contributions, preventing numerical instability.

    \item[] \textbf{LRP-$\alpha\beta$:}
    \[
    R_j = \sum_k \left( \alpha \cdot \frac{a_j w_{jk}^+}{\sum_{j'} a_{j'} w_{j'k}^+} - \beta \cdot \frac{a_j w_{jk}^-}{\sum_{j'} a_{j'} w_{j'k}^-} \right) R_k
    \]
    This rule balances positive ($w_{jk}^+$) and negative ($w_{jk}^-$) contributions to relevance.
    It operates under the constraints $\alpha - \beta = 1$ and $\beta \geq 0$.
    The $\alpha$ term promotes positive contributions, while the $\beta$ term considers negative influences.
\end{itemize}

\label{multiopt} 
\subsection{Multi-objective Optimization Converges to Single-objective Results}

To determine trade-offs between competing interpretability metrics, we used a bi-objective optimization scheme.
This scheme employed a Pareto-based variant of CMA-ES.
It fits a Pareto front across non-dominated solutions, rather than optimizing for a single metric.
The most balanced candidate is selected based on curvature along this front.
Table~\ref{multiobj_results} shows top-performing models for three metric pairings:
Faithfulness vs. Sparseness;
Faithfulness vs. Average Sensitivity;
Sparseness vs. Average Sensitivity.

\begin{table}[ht] 
\centering
\caption{Bi-objective EVO-LRP optimization results for LRP-$\alpha\beta$. Each column pair shows the best solution from optimizing the specified metric combination. Arrows indicate preferred direction. These results demonstrate that bi-objective optimization often yields solutions comparable to strong single-objective outcomes.}
\label{multiobj_results} 
  \begin{adjustbox}{width=\textwidth}
\begin{tabular}{lcccccc}
\toprule
& \multicolumn{2}{c}{\textbf{Faithfulness/Sparseness}} & \multicolumn{2}{c}{\textbf{Faithfulness/Avg Sens.}} & \multicolumn{2}{c}{\textbf{Sparseness/Avg Sens.}} \\ 
 & Mean & Std & Mean & Std & Mean & Std \\
\midrule
\textbf{Faithfulness$\uparrow$}    & 1.44E+00 & 1.08E-01 & 1.44E+00 & 1.09E-01 & - & - \\
\textbf{Sparseness$\uparrow$}        & 6.72E-01 & 1.69E-02 & - & - & 6.72E-01 & 1.62E-02  \\
\textbf{Avg Sensitivity$\downarrow$}  & - & - & 2.73E-01 & 2.07E-02 & 2.73E-01 & 2.10E-02 \\
\bottomrule
\end{tabular}
\end{adjustbox}
\end{table}

\begin{figure}[htbp]
    \centering
    \includegraphics[width=0.75\textwidth]{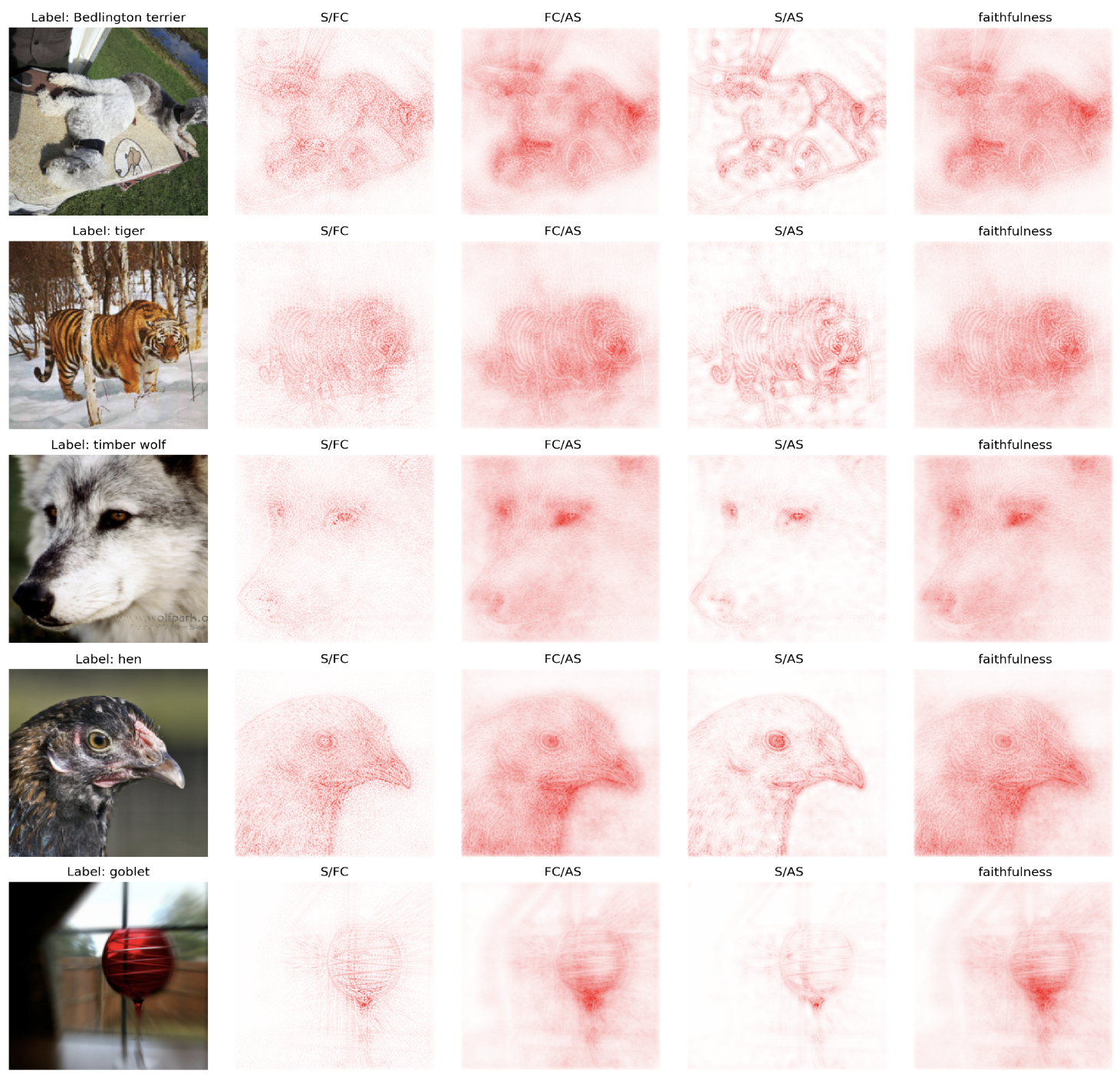}
    \caption{Relevance maps from LRP-$\alpha\beta$ all-class models after bi-objective EVO-LRP optimization for combinations of faithfulness, average sensitivity, and sparseness. For visual clarity and to emphasize the primary positive contributions supporting class predictions, these maps focus on positive relevance scores (red). The rightmost column shows maps from single-objective faithfulness optimization for comparison. The visual similarity across strategies, particularly for specific metric combinations like Sparseness/Faithfulness, supports the quantitative finding that multi-objective optimization often converges to solutions similar to strong single-objective results.}
    \label{fig:URO_mo_metrics} 
\end{figure}

Strikingly, top-performing bi-objective models produced metric values nearly identical to our single-objective optimizations.
This suggests the search landscape structure favors convergence to high-quality solutions, regardless of strategy.
EVO-LRP thus appears robust to the choice between single and multi-objective formulations for these metrics.
Figure~\ref{fig:URO_mo_metrics} illustrates the qualitative similarity of resulting heatmaps.

The Sparseness/Faithfulness and Sparseness/Average Sensitivity combinations yielded visually compelling maps.
These maps effectively highlighted important features with clarity and detail.
This aligns with findings in Section \ref{singl_opt}.
The Faithfulness/Average Sensitivity combination converged to a map almost identical to its single-objective counterpart.
This further supports the robustness of EVO-LRP's optimization process.

\subsection{Additional Qualitative Figures}
\begin{figure}[htbp]
    \centering
    \includegraphics[width=0.75\textwidth]{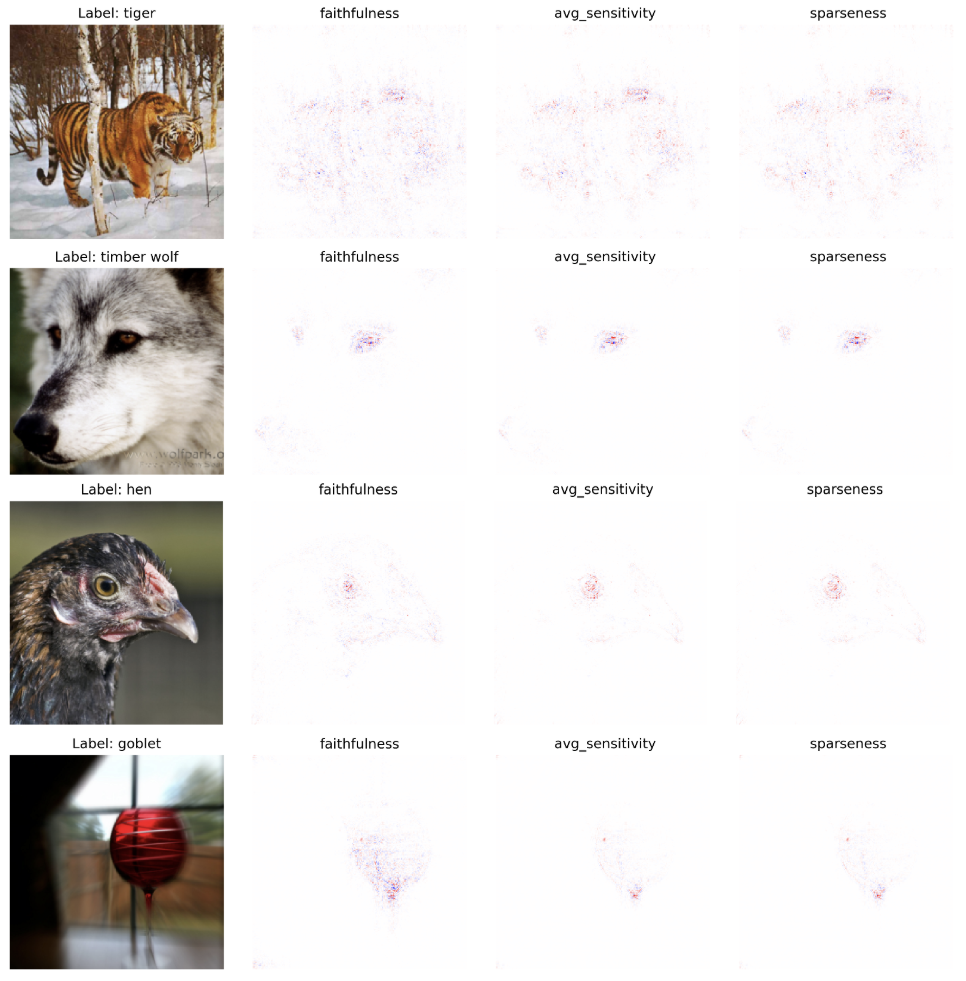}
    \caption{Relevance maps from LRP-$\epsilon$ models after EVO-LRP optimization, targeting faithfulness, average sensitivity, and sparseness individually. The heatmaps are visually similar across all three optimization targets. This similarity empirically confirms the LRP-$\epsilon$ rule's stability and its tendency to produce sparse, consistent attributions with less variation from hyperparameter tuning compared to more adaptable rules like LRP-$\alpha\beta$.}
    \label{fig:UROmetrics_eps} 
\end{figure}

We also ran single-objective experiments on the LRP-$\epsilon$ rule.
As alluded to in Section \ref{limitations}, these results empirically confirm this rule's inherent stability.
They also demonstrate its tendency toward sparse, conservative attributions.
Figure~\ref{fig:UROmetrics_eps} displays these findings. 

\subsection{Broader Societal Impact}
More interpretable and reliable XAI, like EVO-LRP, has many potential societal impacts.
\paragraph{Positive Impacts}
Enhanced transparency can foster trust and AI adoption.
This is key in high-stakes domains like medicine, finance, and autonomous systems.
Revealing model feature reliance aids debugging and identifying biases (e.g., protected attribute use).
This can promote fairer AI.
Clearer explanations can also support scientific discovery. EVO-LRP contributes by identifying data features or complex patterns that human experts might overlook or deem counter-intuitive.
\paragraph{Negative Impacts and Risks}
Over-reliance on explanations is a risk if they are imperfectly aligned with model reasoning or "gamed."
This could lead to misplaced confidence in flawed models.
Poorly interpreted or context-lacking explanations might justify problematic behavior.
Deeper understanding of model vulnerabilities from XAI could inform adversarial attacks (though less direct for LRP).
Computational costs might limit accessibility to well-resourced entities, widening the AI capability gap.
Ethical issues may arise if explanations reveal sensitive data patterns or group vulnerabilities.
\paragraph{Mitigation and Considerations}
Continued research into robust, human-aligned XAI evaluation metrics is crucial.
Human-in-the-loop validation and standardized guidelines for responsible XAI deployment can mitigate risks.
Improving optimization efficiency aids accessibility.
XAI tools like EVO-LRP should augment human oversight, not replace it.

\subsection{Implementation Details}
\label{Compute Resources} 
\subsubsection{Compute Resources}
All experiments can run on a single NVIDIA RTX A6000 GPU.
Increasing algorithm scale (e.g., larger batch sizes) may require more compute resources.
\begin{itemize}
    \item GPU: NVIDIA RTX A6000, 48GB VRAM
    \item CPU: 40 CPUs, x86-64 architecture
\end{itemize}

\subsubsection{Experimental Setup}
Results in Section 4 used EVO-LRP with these hyperparameters for each Uniform-LRP configuration:
\begin{itemize}
    \item Population size: 100
    \item Maximum iterations: 300
    \item LRP-$\alpha$ range: [1, $\infty$] (Note: we constrained $\beta=\alpha-1$)
    \item LRP-$\epsilon$ range: (0, 1]
    \item CMA-ES $\sigma$ (initial step size): 0.5
    \item CMA-ES $\mu$ (mean vector): Initialized at 0 or rule-specific defaults
    \item Batch size: 64
\end{itemize}
CMA-ES population size, $\sigma$, and other internal strategy parameters generally do not require tedious user tuning.
Finding good default strategy parameters is considered part of the CMA-ES algorithm design.

\subsection{Licenses}
The assets we used were as follows:
\begin{itemize}
    \item ILSVRC 2012 ImageNet \cite{russakovsky2015imagenetlargescalevisual}: \url{https://www.image-net.org/challenges/LSVRC/2012}, BSD 3-Clause License Copyright (c) 2017
    \item \texttt{pycma} \cite{hansen2023cmaevolutionstrategytutorial}: The BSD 3-Clause License Copyright (c) 2014 Inria
    \item \texttt{pycomocma} \cite{Tour__2019}: The BSD 3-Clause License Copyright (c) 2019 Inria
    \item Captum \cite{kokhlikyan2020captumunifiedgenericmodel}: \url{https://github.com/pytorch/captum}, BSD-3-Clause license Copyright (c) 2019, PyTorch team
\end{itemize}

\end{document}